# Learning to Combine the Modalities of Language and Video for Temporal Moment Localization

Jungkyoo Shin[a,b] and Jinyoung Moon[a,b,*]

[a]*ICT Major, ETRI School, University of Science and Technology, Daejeon, Republic of Korea ,*
*218 Gajeong-ro, Yuseong-gu, Daejeon, 34129, Republic of Korea,*
[b]*Artificial Intelligent Research Laboratory, Electronics and Telecommunications Research Institute,*
*218 Gajeong-ro, Yuseong-gu, Daejeon, 34129, Republic of Korea*

**Abstract**

Temporal moment localization aims to retrieve the best video segment matching a moment specified by a query. The existing methods generate the visual and semantic embeddings independently and fuse them without full consideration of the long-term temporal relationship between them. To address these shortcomings, we introduce a novel recurrent unit, cross-modal long short-term memory (CM-LSTM), by mimicking the human cognitive process of localizing temporal moments that focuses on the part of a video segment related to the part of a query, and accumulates the contextual information across the entire video recurrently. In addition, we devise a two-stream attention mechanism for both attended and unattended video features by the input query to prevent necessary visual information from being neglected. To obtain more precise boundaries, we propose a two-stream attentive cross-modal interaction network (TACI) that generates two 2D proposal maps obtained globally from the integrated contextual features, which are generated by using CM-LSTM, and locally from boundary score sequences and then combines them into a final 2D map in an end-to-end manner. On the TML benchmark dataset, ActivityNet-Captions, the TACI outperform state-of-the-art TML methods with R@1 of 45.50% and 27.23% for IoU@0.5 and IoU@0.7, respectively. In addition, we show that the revised state-of-the-arts methods by replacing the original LSTM with our CM-LSTM achieve performance gains.

*Keywords:* Temporal moment localization; temporal video localization; temporal video grounding; cross-modal integration; boundary alignment

## 1. Introduction

Due to the great increase in online video consumption in recent years [1], video users require methods for temporal moment localization that are user friendly in order to view interesting video segments that they want to watch among large volumes of videos. Most studies on temporal localization for understanding real-world untrimmed videos have dealt with temporal action localization, which aims to generate action proposals with precise temporal boundaries, including start and end times that cover the ground-truth action instances with high recall. However, the existing methods for temporal action localization and detection [2-8] are not sufficient for video users to identify a specific temporal video segment since predefined action labels have limited representation power for a video segment. To address this, methods for temporal moment localization, which aim to retrieve the best-matching video segment in an untrimmed video corresponding to a moment specified by a natural language query, were introduced in [12-13]. A moment is described by a sentence mainly consisting of nouns, adjectives, verbs, adverbs, and conjunctions. Nouns (with adjectives) and verbs (with adverbs) primarily correspond to objects and their actions in videos, respectively. In contrast with actions, which belong to predefined action categories and occur multiple times in an untrimmed video, there is an assumption that a moment specified by a free-form text query occurs only once in an untrimmed video.

Temporal moment localization is a more challenging problem in localization tasks for untrimmed videos due to the complexity and flexibility of free-from sentence queries for video segments. First, some sentence queries that are temporally located at similar intervals within a video can describe different spatial parts, including different objects and their actions, as shown in Fig 1 (a). Second, a sentence query can describe both its primary submoment and its contextual submoments, as shown in Fig 1 (b), which is located at a temporal interval including the primary moment and excluding its contextual moments. The primary moment is conditioned by the occurrence of its contextual moments, but its temporal interval includes only the primary moment.

* Corresponding author. Tel.: +82-42-860-6712; fax: +82-42-860-6645.
*E-mail address:* jymoon@etri.re.kr

Most existing methods for temporal moment localization generate multiple moment proposals through a sliding-window approach [13-15], single-pass long short-term memory (LSTM) approach [18], anchor-based approach [19-21], score-based frame grouping approach [22], or 2D map-based sampling approach [23] In short, they then project the visual embeddings of the video segments and the query embeddings into a common latent vector space [13, 14, 22, 23] by a specific operation or combination of operations between them, such as concatenation, element-wise addition and multiplication, and average pooling. Most of them evaluate the moment proposals by binary classification

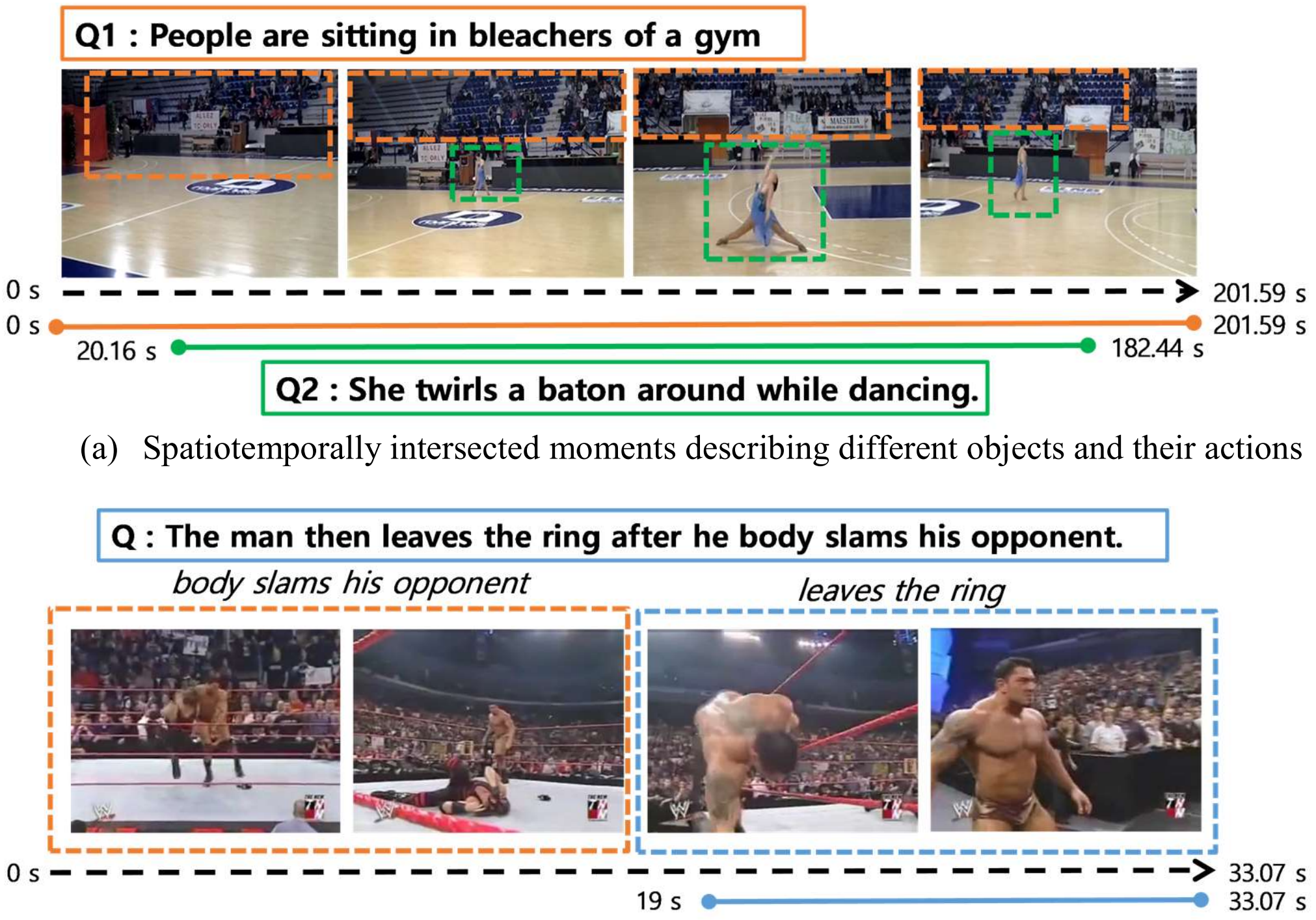

(a) Spatiotemporally intersected moments describing different objects and their actions

(b) A primary submoment conditioned by its contextual submoment

Fig. 1. Challenges of temporal moment localization through free-form sentence queries

with the fused embedding vectors of the moments. Some methods generate the fused embedding vector of a whole video and then directly predict the start and end times of the moment without proposal generation [24, 25].

To provide a natural way to fuse the multimodal information of a moment, existing methods generate fused contextual features by combining query and video embeddings that are generated independently. Some methods use attended query embeddings [15, 18] and other methods use attended visual embeddings for a whole video [20, 21, 25, 26]. However, attention-based methods have certain limitations. First, they overlook that a sentence query for a moment can describe not a full frame but a partial 3D cube in a video segment, as shown in Fig. 1 (a). Although some methods generate the attended visual embeddings, the visual embeddings corresponding to a full-frame video segment are attended with an attention score. Therefore, they are not able to focus on partial 3D cubes in a video segment. Second, many methods do not apply attention mechanisms for query embeddings, which are essential for complex queries. A complex query consists of multiple subqueries for a central moment, which determines its temporal interval, and for contextual moments, which are located before or after the central moment, as shown in Fig. 1 (b). Because the visual embeddings corresponding to a clip represent a specific submoment related to a subquery, an attention mechanism for query embeddings is fundamental for determining precise moment boundaries. In addition, several existing methods that fuse query embeddings and visual embeddings for a moment or a whole video at once [13, 14, 22, 24, 25] fail to consider visual information accumulated from contextual moments, which is excluded from the moment boundaries. Additionally, other methods accumulate such information using a traditional LSTM after concatenating both embeddings [15, 18, 19]. Finally, attention-based methods assume that the use of attended visual embeddings is effective in localizing precise moment boundaries, but it can actually lower the localization performance by ignoring necessary visual information.

Moreover, score-based methods for temporal moment localization utilize the boundary-aware scores for generating moment proposals or for obtaining the final confidence scores for moment proposals in traditional ways, which are adopted in existing methods for temporal action localization. The semantic activity proposal (SAP) [22] predicts visual-semantic correlation scores and generates the initial moment proposals by grouping consecutive frames with predicted high confidence scores above a specific threshold, which is similar to temporal actionness grouping (TAG) in [3] and adds extended moment proposals in a sliding-window approach within each initial moment proposal.

To address these limitations, we propose a two-stream attentive cross-modal interaction network (TACI) for temporal moment localization based on our cross-modal LSTM (CM-LSTM), which accumulates contextual visual and textual information to obtain integrated contextual features and a temporal boundary-aware network that obtains a 2D confidence score map for all possible moment proposals in the end. To mimic the cognitive process for temporal moment localization by which people consistently remember a query while watching a video, focus on a part of the video segment related to the part of the query, and gather contextual information related to the query throughout the entire video, we reformulate the computations of the existing gates and add a modal gate to the original LSTM unit to mutually combine modalities by taking both the visual and query embeddings as input. Specifically, we generate a cross-modal filter that is responsible for the relationship between the two modal inputs and then fuse two modal inputs based on a cross-modal filter and modal gate. The modal gate controls the importance of the information contained in the query and video segment from one time to the next. This structure not only selects the query-related video features but also selects the video-related query features, which enables our model to understand more complex and detailed multimodal information.

Moreover, inspired by [27], we designed a two-stream attention mechanism for modal interaction. As the attention layer encourages the model to focus on a certain location of input, mandatory information may exist on other locations. To prevent necessary visual information from being ignored, we introduce two attention streams for attended and unattended visual embeddings and generate two fused cross-modal embeddings individually.

Furthermore, we propose a boundary-aware network that predicts three boundary-related score sequences for the start, end, and momentness using the integrated contextual features; we obtain the final 2D confidence score map by using both the global confidence score map from the convolutional layers that take the integrated contextual features as input and the local confidence score map from the convolutional layers that take the predicted boundary-aware score sequences as input.

To summarize, the main contributions of our paper are as follows:

- We introduce a novel recurrent unit, CM-LSTM, that both fuses visual and query features complimentarily and encodes the spatiotemporal contextual features with a modality combination mechanism. In contrast with existing methods, the TACI with CM-LSTM integrates the two modalities and generates the long-term contextual features at the same time.
- We devise a novel two-stream attention mechanism for modal interaction, which takes in both attended and unattended visual memory and generates two contextual features for late-fusion.
- We propose a TACI that generates a final 2D moment proposal map by combining the two maps obtained locally from the three predicted boundary-aware score sequences as well as the integrated contextual features obtained globally from CM-LSTM in an end-to-end manner.
- We demonstrate that our TACI outperforms state-of-the-art methods in extensive experiments on the major benchmark dataset, ActivityNet-Captions. In particular, the TACI shows the best performance for complex sentence queries, including submoments with both spatial and temporal transitions, compared to the two state-of-the-art methods. In addition, we show that CM-LSTM brings performance improvements when applied to three recent methods that use the original LSTM.

## 2. Related Work

We review previous studies related to our approach, which are categorized into three research areas: (1) action recognition in trimmed videos, (2) temporal action localization and detection in untrimmed videos, and (3) temporal moment localization in untrimmed videos with sentence queries.

### *2.1. Action Recognition in Well-Trimmed Videos*

Action recognition is a fundamental area in video understanding. Given a well-trimmed video, the goal of action recognition is to recognize the action class of an action instance included in the input video. Similar to object recognition and detection, deep learning methods have significantly improved the performance of action recognition compared to early works using handcrafted features. Some video back-bone networks have been proposed for action recognition [27, 28, 29, 9]. The two-stream network for action recognition [27], which was the first action recognition model that surpassed traditional models using hand-crafted features, extracts appearance and motion information from an RGB frame and optical flow frames, respectively, and then combines them through late fusion. The C3D network [28] feeds 16 consecutive frames to 3D convolutional networks to extract appearance and motion information directly from raw RGB frames. The I3D network [29] bundles the RGB and optical flows in 64 frames and adopts 3D convolutional networks to better learn the appearance and motion features simultaneously. The publicized models, including the two-stream, C3D, and I3D networks pretrained on Sports-1M [30], and Kinetics [29], are widely employed as backbone networks to extract unit-level video clip features in various video understanding areas, such as video classification, temporal action localization and detection, and temporal moment localization. Recently, [9] proposed a new attention network that constructs video representations with two-level attention to learn local-global information to achieve comparable performance with using only RGB information.

### *2.2. Temporal Action Localization and Detection in Untrimmed Videos*

The aim of temporal action localization is to obtain accurate temporal intervals representing action in-stances within a given untrimmed video. Methods for temporal action localization are required to under-stand real-world untrimmed videos, primarily including numerous backgrounds and some meaningful action instances that belong to multiple action classes. Temporal action detection is composed of temporal action localization and action recognition for localized action proposals. Existing temporal action localization methods are divided into supervised methods with full information of action instances, including action class and temporal intervals [3, 4, 5].

Inspired by successful object detection approaches [34, 35] consisting of two stages—proposal generation and classification—the temporal action detection process in most works also consists of two steps, generating proposals and predicting proposal scores based on class-agnostic actionness. Zhao et al. [3] suggested a temporal action proposal generation scheme, TAG, which has been widely utilized in temporal action detection methods based on actionness scores to provide action proposals that are more promising than sliding window-based action proposals.

To overcome the limitations of the methods using a sliding-window scheme, score-based methods have been proposed for temporal action localization. Lin et al. [4] proposed the boundary-sensitive network (BSN), which was the first temporal localization method that showed high performance in a completely score-based manner, moving away from the sliding-window approach. The BSN [4] generates score-based temporal action proposals and predicts proposal-level confidence scores using the proposal features corresponding to the proposal intervals. The BSN [4] was extended to the boundary-matching network (BMN) [5] using a predicted 2D map for the proposal confidence scores.

### *2.3. Temporal Moment Localization by Sentence Queries*

Temporal moment localization is a relatively new task proposed by [12] and [13], the aim of which is to find the temporal boundaries with start and end times within the video that best match a given sentence query.

Earlier methods for temporal moment localization [12, 13, 22] used basic computational methods such as concatenation, element-wise addition, and element-wise multiplication to estimate the relationship between video segments and queries. [12] concatenated the segment feature with the entire video feature as a single moment representation to exploit both the local and global action features. Both the moment and sentence features are projected into the same vector space. The squared distance between the two projected vectors is assumed to approximate the alignment scores for the video segments. [13] encoded the visual features of a video segment, including a central clip and the pre-context and post-context clips that correspond to a moment proposal as well as sentence features, using Skip-thoughts or LSTM. They proposed the cross-modal temporal regression localizer (CTRL) to compute alignment scores for moment proposals. CTRL consists of three concatenated features and fuses video and sentence embeddings by element-wise addition, element-wise multiplication, and vector concatenation followed by a fully connected layer. [22] computed the alignment score for every unit-level video clip and sentence by element-wise multiplication and generated a proposal by grouping the clips with a high alignment score. However, calculating the relationship between a video segment and a given sentence using basic computational methods has limitations. It is difficult to understand the complex relationship between the video and the sentence because such methods consider only regional relations and do not take contextual information into account.

To address this problem, most of the recent studies on temporal moment localization are applied to various types of attention modules to determine complex contextual information, similar to multi-modal attention studies for other tasks, including video captioning [10] and video QA [11], which generate language-related video features and video-related language features.

Some methods [20, 25, 26] attend to the whole video at once based on the sentence features. [26] reweighted the video features by temporal attention weights. Instead of explicitly using attention, the inner product between the sentence features and the video features was used as the temporal attention weight. [25] directly predicted the probability of each video clip being the start or end point of the video segment that matches the sentence. Temporal moment localization with guided attention (TMLGA) is guided by a dynamic filter that is based on a principal attention mechanism that encourages the model to focus on the sentence-related video clips. [20] employed dynamic filters to emphasize sentence-related video features, generated temporal proposals over different temporal locations and scales via a hierarchical convolutional network and explicitly modeled temporal relationships among different proposals in a graph-structured network. Recently, [37] proposed a local-global network, which divides the query feature into multiple semantic entities and reflects bi-modal interactions between the linguistic features of the query and the visual features of the video in multiple levels. However, these methods have some limitations. The video-level visual features are given as input to the attention layer, which generates a dynamic filter in a global perspective for the whole video. This can suppress some local information that includes crucial context.

To distinguish sentence-related information in each frame without losing local information, some methods for temporal moment localization [18, 21] apply a frame-specific attention module. [18] proposed Match-LSTM, which is composed of three LSTM layers: the first layer is for textual information, the second layer is for visual information, and the last layer is for boundary alignment. Each video frame is attentively matched to a specific word in the sentence. [21] attached sentence features to each video clip and generated scaling and shifting parameters. The two modulation parameters adjusted each video clip to associate and compose the sentence-related video clips. [17] divides the video feature into multiple temporary scales to interact with the given query. Based on this interaction, a hierarchical context-aware 2D correlation map is generated to model the relation.

However, because a video and a sentence have very complex structures, the video clips may not fully correspond to the sentence information. Because the video clips in the ground-truth moment may contain only part of the sentence in-formation, it is possible to misjudge the matching clip.

Because of these problems, understanding both the video and sentence by cross-modal interaction is required. Recent methods for temporal moment localization [24, 19] applied cross-modal attention to a query and video. [24] proposed a multimodal co-attention mechanism that uses an attention module on both the video and sentence. Video attention exploits the relative associations between different video clips and queries, which can reflect the global temporal structure of the video. Query attention weights the video-related details in word sequences to provide clear guidance for temporal local prediction. [19] applied an attention layer to aggregate the sentence features of each frame. Then, cross-gate attention was applied to the aggregated query representation and original video feature to emphasize the crucial content and weaken the inessential parts of both features. [16] models the modality interaction in both the sequence level of a given query and segment level of a given video in a pairwise fashion. However, the module is responsible for learning contextual information within the video using only video features without sentence features. Since a video contains various complex contextual information, this method may fail to extract the contextual information that matches the sentence. Additionally, since multiple pieces of information may exist in a single video segment, the model can match sentences with different contexts.

[23] generated a two-dimensional temporal feature map in which the dimensions represent the start and end clip indexes. The features of the video segments corresponding to each start and end index are pooled and stacked as a 2D proposal map. Zhang et al. stacked convolutional layers onto the 2D proposal map after element-wise multiplication between visual and query features to perceive complex moment contextual information. 2D-TAN [23] is similar to our TACI in the sense that it generates a 2D proposal map to localize video. However, our approach is different from 2D-TAN in understanding and modeling relationship between a video and query. As 2D-TAN divides a video feature into several moment proposal features corresponding to a 2D map and applies a convolution network to encode the adjacent temporal relation between proposals. Since this map interacts with the feature of every proposal possible, applying a convolutional network towards a two-dimensional temporal feature map requires a vast amount of computational overhead. In contrast, we perform an effective cross-modal interaction to model correlations between video and query, obtain contextual integrated feature, and then directly generated a 2D proposal map from the contextual integrated feature.

Most of the methods introduced above train the model using only the momentness and then globally extract the proposal scores for video segments from the entire video. For this reason, the model lacks information on the temporal boundaries, which can lead to inaccurate predictions. To solve this problem, [18] predicted the probability of each clip being the boundary. Then, proposals were selected to be boundary-aware by adding boundary possibilities corresponding to the start and end points of the predefined proposal with the proposal score. This gave the model guidance to reliably distinguish between the matching video information and the nonmatching video information. However, the probabilities of the boundaries did not distinguish between the start point and end points, which could cause confusion regarding which of the distinct sides of the video matches the sentence. [25] calculated only the probabilities of a pair of start and end points of a temporal moment for each video clip. However, this calculation was learned only from a local perspective.

## 3. Proposed Method

We propose a novel approach for temporal moment localization by sentence query. The key idea is to accumulate contextual information from both visual and semantic embeddings using CM-LSTM and obtain confidence scores for densely sampled moment proposals in the 2D proposal map obtained globally from the integrated information and locally from the predicted boundary-aware scores, as shown in Fig. 2.

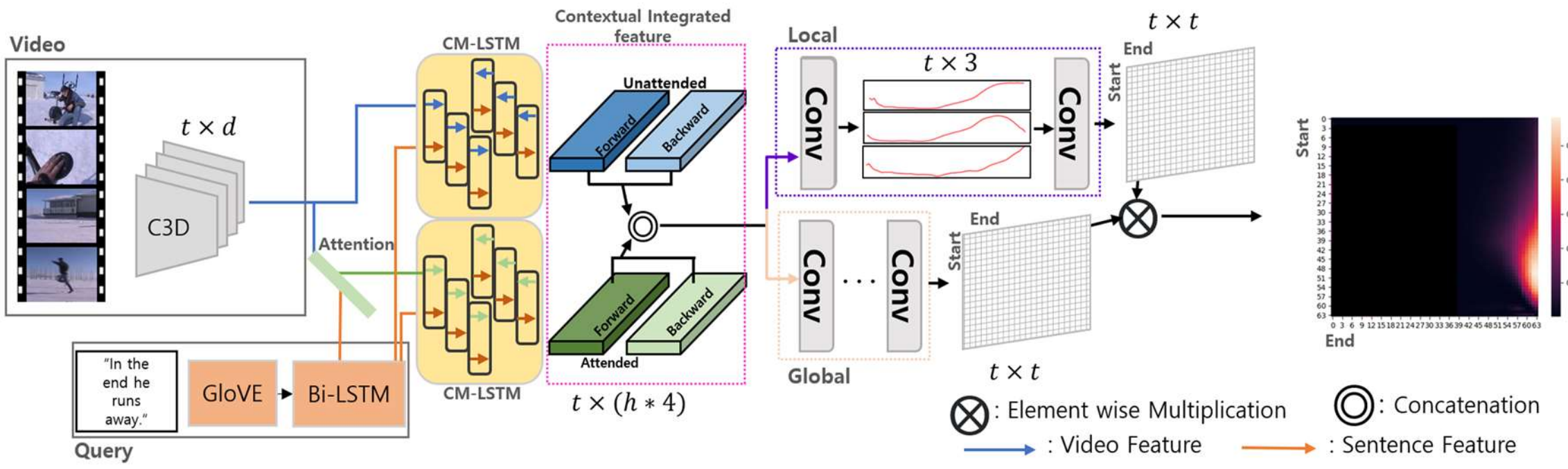

Fig. 2 shows the overall architecture of our algorithm, which consists of a video and query encoder (Section 3.2), two-stream attention (Section 3.3) and CM-LSTM (Section 3.4), to obtain the contextual integrated features. We generate a 2D proposal map and evaluate each cell in a local-global manner (Section 3.5) to obtain the best matching video proposal.

### *3.1. Problem Formulation*

Given an untrimmed video V and a sentence query $S = \{w_q\}_{q=1}^{N}$, where T is the length of the input video and N is the number of words in the sentence, the goal of temporal moment localization is to locate the temporal boundaries of the moment described by the sentence query, namely, $(t_{start}, t_{end})$.

### *3.2. Encoders*

**Video Encoding.** We first extract the unit-level clip features for all video clips. To extract denser features and to unify the sizes of the unit-level clip and sentence features, a fixed number of clip features are fed to the fully connected layer. The representation of a video feature is denoted as $F^v = \{f_1^v \cdots f_l^v\} \in \mathbb{R}^{dxl}$, where $f_t^v$ is the unit-level clip feature at the t-th clip and l is the fixed number of clips for an input video.

**Sentence Encoding.** Given a sentence query $S = \{w_q\}_{q=1}^{N}$, where wq is the q-th word and N is the number of words in the sentence, we employ a GloVe Word2Vec model pretrained on Common Crawl [31] to convert each word to word embedding vectors without fine-tuning. Specifically, the sentence query S made up of N words is embedded in $Q \in \mathbb{R}^{300\times N}$ by the GloVe model. Then, the word embeddings Q are sequentially fed to the bidirectional LSTM network. The last hidden states of the bidirectional LSTM are used as the sentence representation. We unify the size of both the sentence and the unit-level clip features as d. The final sentence feature is denoted as $f^q \in \mathbb{R}^{dx1}$, where d is the size of the dimension of the hidden state of the bidirectional LSTM network.

### 3.3. Two-Stream Attention

To emphasize specific clip features in the video that are related to a sentence, we attend to the entire set of video features $F^v$ with respect to the entire set of encoded sentence features $f^q$. The attention mechanism for the video features is formulated as follows:

$$H = tanh(W_b F^v + (U_b f^q)\mathbf{1}^T + b\mathbf{1}^T) \in \mathbb{R}^{k\times l}, \quad (1)$$

$$h^a = softmax(W_a H) = \{a_1, \dots, a_l\} \in \mathbb{R}^{l}, \quad (2)$$

$$f_t^a = h^a * f_t^v, \quad (3)$$

$$F^A = \{f_1^a, \ldots, f_l^a\} \quad (4)$$

where $W_b, U_b \in \mathbb{R}^{k\times d}$ and $W_a, b \in \mathbb{R}^k$ represent the learnable parameters, k is the dimension of the hidden states of the attention module, $\mathbf{1}^T$ is the transpose of an all-ones vector, and $\otimes$ is element-wise multiplication. Through the attention mechanism, we suppress the irrelevant features and pass only the necessary information to the next step.

Our attention-based early fusion method is efficient but can neglect relations between clips within a whole video, which can lead to the suppression of clip features containing contextual information. To address this limitation, we propose a two-stream attention structure consisting of one stream with attention and another without attention. We pass two kinds of video features, the attended video feature $F^A \in \mathbb{R}^{d\times l}$ and the raw video feature $F^v \in \mathbb{R}^{d\times l}$, to the following bidirectional CM-LSTM.

3.4. Modality Fusion Using Cross-Modal LSTM

A single video clip that matches a query may contain only a portion of the query and may also contain information other than the query's activity at the same time. Therefore, it is necessary to selectively learn only the pieces of information that are related to each other. For temporal moment localization for a given sentence query, people constantly recall the sentence while watching the input video, focus on the relevant content, and accumulate the contextual information from the past to the present. By mimicking this cognitive process of humans, we propose a new recurrent network based on the LSTM structure that selectively uses the video clip feature related to the query and accumulates the contextual information.

**Original LSTM.** As shown in Fig. 3, LSTM enables temporal information to flow along cell state $C_t$ by removing or adding information related to the cell state that is regulated by three gates: forget gate, input gate, and output gate.

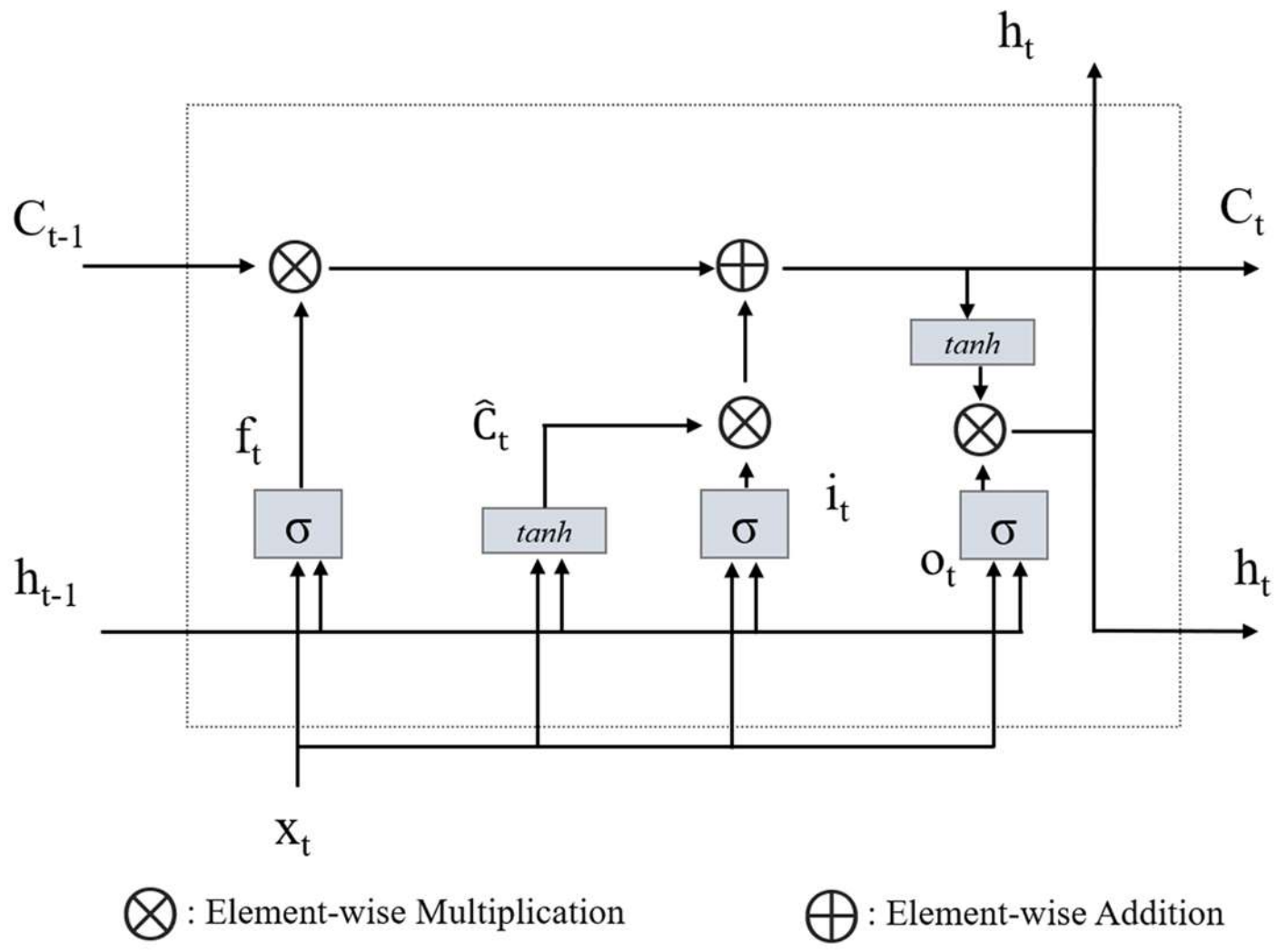

Fig. 3. Original LSTM unit

$$f_t = \sigma(W_f x_t + U_f h_{t-1} + b_f), \quad (6)$$

$$i_t = \sigma(W_i x_t + U_i h_{t-1} + b_i), \quad (7)$$

$$\widehat{C}_t = tanh(W_c x_t + U_c h_{t-1} + b_c), \tag{8}$$

$$C_t = f_t \otimes C_{t-1} + i_t \otimes \tilde{C}_t, \tag{9}$$

$$o_t = \sigma(W_o x_t + U_o h_{t-1} + b_o), \tag{10}$$

$$h_t = o_t \otimes tanh(C_t), \tag{11}$$

where $\otimes$ is element-wise multiplication and σ is the sigmoid function. For a given input $x_t \in \mathbb{R}^d$ at the t-th time step, $W_f, W_i, W_o, W_c \in \mathbb{R}^{p\times d}$ , $U_f, U_i, U_o, U_c \in \mathbb{R}^{p\times p}$ , and $b_f, b_i, b_o, b_c \in \mathbb{R}^p$ are learnable parameters and p is the number of hidden states. We obtain the new cell state $C_t$ by using the old cell state $C_t$ and the new vector $\widehat{C}_t$, which are controlled by the forget $f_t$ and input $i_t$ gates, respectively. The hidden state $h_t$, which represents the contextual information summarized from the initial state to the current state, is determined by the cell state $C_t$ regulated by the output gate $o_t$.

**Our CM-LSTM.** We propose a new recurrent model, CM-LSTM, which is based on LSTM and simultaneously calculates the modal dependencies between visual input sequences and semantic input sequences, as shown in Fig. 4. In contrast to the unit of the original LSTM, which takes only the single-modal input $x_t$, our CM-LSTM unit takes the clip feature $f_t^v$ and the sentence feature $f^q$ at the t-th time step to learn the modal relationship between two input and the contextual information within the video feature simultaneously. Specifically, we first generate the attention filter which interacts the visual clip feature and sentence feature to calculate the relationship between the two modal inputs.

$$a_{filter} = softmax(W_e(tanh(W_c f^q + W_d f_t^v)), \tag{12}$$

where $W_c, W_d, W_e \in \mathbb{R}^{d\times d}$ are learnable parameters. As mentioned earlier, related information within the sentence and video clips may vary in different timesteps. Therefore, to attend to the related information within each feature, we interact visual clip feature and sentence feature with attention filter respectively, as follows

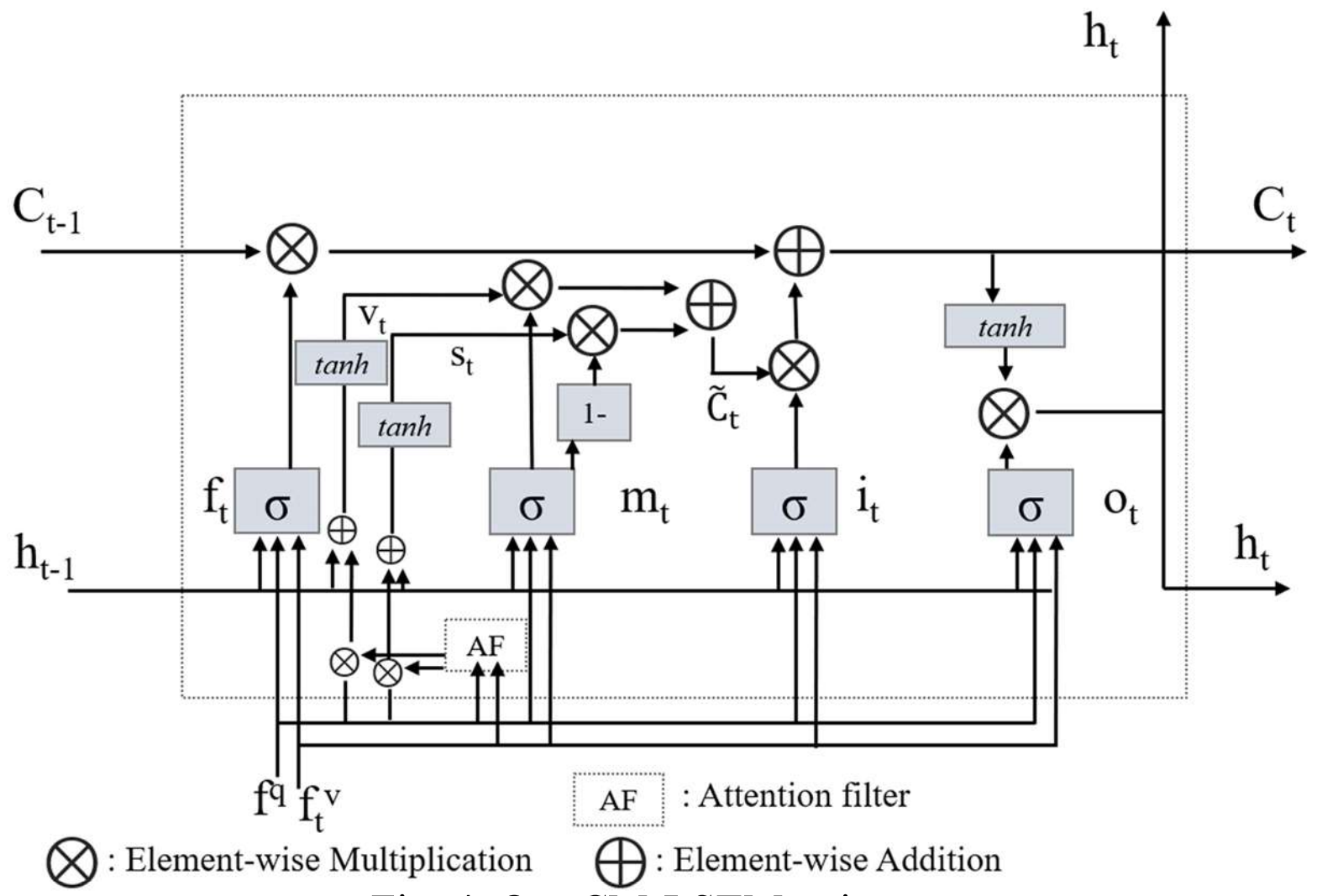

Fig. 4. Our CM-LSTM unit

$$v_t = tanh(a_{filter} \otimes W_v f_t^v + U_c h_{t-1}), \tag{13}$$

$$s_t = tanh(a_{filter} \otimes W_s f^q + U_c h_{t-1}), \tag{14}$$

where $W_v, W_s \in \mathbb{R}^{p \times d}$ and $U_c \in \mathbb{R}^{p \times p}$ are learnable parameters and p represents the number of hidden states.

$v_t$ and $s_t$ represents visual clip and sentence feature attended on spatial modal relation respectively, which is newly obtained for each timestep. To supplement contextual information, we apply the baseline from LSTM and generate forget gate, input gate, and output gate using $v_t$ and $s_t$, and interact the features from the current step with the features from the previous step to encode spatiotemporal contextual feature. We fuse this feature with a newly added gate, called a modal gate $m_t$. At each timestep, the amount of substantial information for modal relation within each sentence and visual clip may differ. The proposed modal gate is responsible for controlling the balance between the attended sentence feature and the attended visual clip feature.

$$f_t = \sigma(W_{fx} f_t^v + W_{fy} f^q + U_f H_{t-1} + b_f), \tag{15}$$

$$i_t = \sigma(W_{ix} f_t^v + W_{iy} f^q + U_i H_{t-1} + b_i), \tag{16}$$

$$m_t = \sigma\left(W_{mx} f_t^v + W_{my} f^q + U_m H_{t-1} + b_m\right), \tag{17}$$

$$\tilde{C}_t = m_t \otimes v_t + (1 - m_t) \otimes s_t, \tag{18}$$

$$C_t = f_t \otimes C_{t-1} + i_t \otimes \tilde{C}_t, \tag{19}$$

$$o_t = \sigma(W_{ox} f_t^v + W_{oy} f^q + U_o h_{t-1} + b_o), \tag{20}$$

$$h_t = o_t \otimes tanh(C_t), \tag{21}$$

where $W_{fx}, W_{ix}, W_{ox}, W_{mx}, W_{fy}, W_{iy}, W_{oy}, W_{my} \in \mathbb{R}^{p \times d}$, $U_f, U_i, U_o, U_m \in \mathbb{R}^{p \times p}$, and $b_f, b_i, b_o, b_m \in \mathbb{R}^p$ are learnable parameters and p is the size of the hidden states. Specifically, we fuse the video clip features and query features with a cross-modal filter and modal gate. The cross-modal filter contains the related spatial information, and the modal gate reweights each feature, fusing only the sentence-related partial video clip features and video-related partial query features. The temporal information is then obtained by an input gate with a forget gate, and with this structure, our proposed unit can exclude the background information and focus solely on the query-related contextual information.

We utilize the recurrent layers based on the CM-LSTM unit to bidirectionally process visual and sentence features for both the attended video features and the unattended video features and concatenate all hidden states to obtain the contextual integrated features $F_i \in \mathbb{R}^{p \times l}$.

## 3.5. Local-Global Proposal Evaluation

We generate a 2D proposal map in which each location (x, y) represents a moment proposal, with x as the start time and y as the end time, not only to evaluate the confidence score of all possible proposals but also to perceive the relationship information through the surrounding proposals. This helps the TACI to better understand the contextual information and choose the best proposal by comparing the amount of matching information each adjacent proposal contains.

We propose a local-global proposal generation module. Globally, our module predicts a 2D confidence score map for moments with the integrated contextual features. Additionally, we extend our proposed evaluation approach with a local proposal module that focuses on the boundary-aware information based on the start, end, and moment scores.

**Global Proposal Evaluation Submodule.** The global evaluation submodule generates a 2D map of the moment confidence scores directly from the integrated contextual features $F_i$. This submodule consists of a stacked multiple-convolution layer conv($\theta_k$, $\theta_s$, $\theta_f$), batch normalization and parametric ReLU. $\theta_k$ is the size of the kernel, $\theta_s$ is the size of the stride and $\theta_f$ is the size of the filter. Batch normalization and parametric ReLU [33] are used to avoid gradient vanishing. Derived from ReLU, parametric ReLU, (pReLU) is a nonlinear activation that updates the negative inputs through a learnable variable gradient.

$$f(x_i) = \begin{cases} x, & \text{if } x_i > 0 \\ a_i x_i, & \text{if } x_i \leq 0 \end{cases} \tag{22}$$

where $a_i$ is a learnable parameter. A sigmoid is applied to the output of the last layer. The moment proposals in the 2D proposal map are extracted directly for the whole video, and each one represents the confidence score for a specific video proposal in the given input video. By stacking multiple layers, we construct a convolutional network that enables us to compare the proposals. We denote the output of this generation as $M_{global} \in \mathbb{R}^{l \times l}$.

**Local Proposal Evaluation Submodule.** The local evaluation submodule takes the same input as the global generation submodule. In contrast to the global evaluation submodule, it generates three boundary scores for the moments locally by using the start, end, and momentness. Similar to the actionness in action localization and detection, which is a class-agnostic indicator of whether a temporal point is included in an action instance, momentness is a class-agnostic indicator of whether a temporal point is included in a moment instance. This gives guidance for the model to clearly distinguish a matched moment from the background. Taking the boundary-aware scores as $L = \{L_{start}, L_{end}, L_{momentness}\} \in \mathbb{R}^3$, the local evaluation submodule ultimately generates a 2D confidence score map of the moments through stacked convolution layers. To generate three 1D boundary-aware scores and a 2D score map, pReLU, batch normalization and a convolution layer are used, and the sigmoid is processed at the last layer they have in common. We denote the output of this generation process as $M_{local} \in \mathbb{R}^{l \times l}$.

**Ensemble.** To fuse the two obtained 2D maps, we apply late fusion by element-wise multiplication. We multiply by a mask to ignore the invalid moment proposals in which the end time precedes the start time.

$$\text{Mask} \in \mathbb{R}^{l \times l} \tag{23}$$

$$\text{Mask}(t_{start}, t_{end}) = \begin{cases} 1, & \text{if } t_{start} \leq t_{end} \\ 0, & \text{if } t_{start} > t_{end} \end{cases} \tag{24}$$

$$M = M_{global} \otimes M_{local} \tag{25}$$

$$M = \text{Mask} \otimes M \tag{26}$$

Then, we derive the proposal corresponding to the cell with the highest score in the 2D Map as the correct answer.

$$T = \text{argmax}(M) \tag{27}$$

Finally, we ground the video by extracting the X-axis and Y-axis of T as the start and end points of the original video.

## 3.6. Training

Our model is trained with two main losses related to the global and local proposal evaluation submodules.

**Map Loss.** To find a moment proposal that best matches our 2D score map of moments, we calculate the intersection-over-union (IoU) scores between each moment proposal and the ground truth ($t_{start}$, $t_{end}$). We set the IoU scores in the 2D score map as the ground truth. We use the binary cross-entropy between the proposed 2D score map and the ground truth. The formula is expressed as follows:

$$\mathrm{Loss_{Map}} = \frac{1}{c} \Sigma_{i=1}^{C} y_i \log p_i + (1 - y_i) \log(1 - p_i) \tag{28}$$

where p is the calculated possibility for each cell within the generated proposal map and y is the IoU value in the ground-truth map.

**Local Loss.** We apply soft labelling for the start and end ground truth $gt_s, gt_e \in \mathbb{R}^l$. For the momentness ground truth, $gt_m \in \mathbb{R}^l$ is set as 1 if the clip is within the action annotation and 0 otherwise. We adopt the Kullback-Leibler divergence loss between the three predicted and ground-truth possibility distributions. The formula is as follows:

$$K(P(i)|(Q(i)) = \Sigma_{i=1}^{C} P(i) \log\left(\frac{P(i)}{Q(i)}\right) \tag{29}$$

where P(i) is the calculated possibility and Q(i) is the ground truth. The boundary loss is the sum of the three individual losses for the start, end, and momentness scores.

$$\mathrm{Loss_{local}} = K(L_{start}|gt_s) + K(L_{end}|gt_{end}) + \lambda_m K(L_{momentness}\,|gt_{momentness}) \tag{30}$$

The final loss is calculated as

$$\mathrm{Loss} = \mathrm{Loss_{Map}} + \lambda_{local} * \mathrm{Loss_{local}} \tag{31}$$

## 4. Experiments

### *4.1. Datasets*

**ActivityNet-Captions.** The ActivityNet-Captions [32] dataset contains 19,209 untrimmed videos that involve activities of daily human life. Each video is two minutes long on average. Following the setting in [19] for the cross-modal interaction network (CMIN), we use val_1 as the validation set and val_2 as the testing set; the whole dataset has 37,417, 17,505, and 17,031 moment annotations for training, validation, and testing, respectively. Currently, this is the largest dataset for temporal moment localization. The details of ActivityNet-Captions are summarized in Table 1.

**Charades-STA.** The Charades-STA [13] dataset contains 9,848 videos that mainly involve indoor activities. The average length of a video is 30 seconds. Charades-STA contains 12,408 and 3,720 moment annotations in the training and testing sets, respectively. The details of Charades-STA are summarized in Table 1.

Table 1. Summary of the ActivityNet-Captions and Charades-STA datasets.

| Dataset | Video | Queries | Average Video Length | Average Query Length | Query Per Video |
|---|---|---|---|---|---|
| ActivityNet-Captions [32] | 19,209 | 71,953 | 120.81 | 13.12 | 3.74 |
| Charades-STA [13] | 9,848 | 16,110 | 31.11 | 6.21 | 1.63 |

### *4.2. Implementation Details*

**Data Preprocessing.** For ActivityNet-Captions, we used the C3D network [23] as a video feature extractor. A C3D feature per every 8 frames was extracted from the fc7 layer of the pretrained C3D network. To reduce the dimension of the C3D feature from 4096 to 500, principal component analysis (PCA) was applied to the C3D features. For Charades-STA, we used VGG features as video representations. Frames were sampled from the video at 6 frames per second (fps), and pretrained VGG-16 was applied to every frame to extract the video features, with 4096 dimensions per frame. We fixed the length of a video feature to 64 through linear interpolation. In a sentence query, words were converted to a pretrained GloVe word embedding vectors.

**Model Setting.** For ActivityNet-Captions, we set both the hidden size of bi-LSTM for sentences and CM-LSTM to 256. To avoid overfitting, we used two dropout layers in two recurrent layers and set the dropout probability to 0.5. We stacked two CNN layers for global proposal evaluation, two CNN layers for generating the local possibility and four CNN layers for local proposal evaluation. The number of hid-den states for attention was set to 32. For Charades-STA, we set the hidden size of bi-LSTM for sentences and CM-LSTM to 128 and 256, respectively. The probability for the dropout layers in bi-LSTM and CM-LSTM was set to 0.5 and 0.2, respectively. We stacked four CNN layers for global proposal evaluation, score sequence prediction, and local proposal evaluation.

**Optimization.** We used Adam with a learning rate of 0.001 and a batch size of 32. The parameters $\lambda_{local}$ and $\lambda_m$ were set to 2 by grid search.

### *4.3. Evaluation Metric*

Following the method used in a previous studies [12, 13], we adopt "R@n, IoU@m" as our evaluation metric. "R@n, IoU@m" describes the percentage of top-n retrieved proposals that have at least one temporal IoU value larger than m. To calculate R@n, IoU@m, we first compute the IoU between the selected moment proposal and the ground-truth moment. Then, to evaluate the overall performance, we divide the number of top-n selected queries that have IoU values higher than m by the total number of queries. We denote this evaluation as $\mathrm{R(n,m)} = \frac{1}{\mathrm{N_q}}\sum_{\mathrm{i=1}}^{\mathrm{N_q}} \mathrm{f(n,m,q_i)}$, where R(n, m) means "R@n, IoU@m", $N_q$ is the number of queries and $\mathrm{f(n,m,q_i)} = \begin{cases} 1, & \text{if Top} - \text{n IoU}(\mathrm{q_i}) \geq \mathrm{m} \\ 0, & \text{if Top} - \text{n IoU}(\mathrm{q_i}) < m \end{cases}$. We denote R@1, IoU@0.3, R@1, IoU@0.5, R@1, IoU@0.7 performance of our model for evaluation.

### *4.4. Comparison with Other Methods*

**Performance.** We evaluate the performance of our proposed network on two datasets as a benchmark and compare it against the following methods:

- Simple computational fusion. We compare our approach to the moment context network (MCN) [12], CTRL [13], SAP [22], Contextual Pyramid Network, and 2D temporal adjacent network (2D-TAN) [23], which fuse the query and video features by simple computations, such as concatenation and element-wise multiplication. These methods do not rely on attention layers.

- Video-wise attention fusion. We compare our approach to the moment alignment network (MAN) [20], query-guided segment proposal network (QSPN) [26] and TMLGA [25], which fuse the query and video features with an attention layer that attends to the whole video feature at once and local-global interaction (LGI) [36].
- Clip-wise attention fusion. We compare our approach to compact bilinear pooling (CBP) [18], semantic conditioned dynamic modulation (SCDM) [21], pairwise modality interaction (PMI) [16] and contextual pyramid network (CPNet) [17] which fuse the query and video features with an attention layer that attends to one clip feature at a time.
- Cross-modal attention fusion. We compare our approach to attention-based location regression (ABLR) [24] and CMIN [19], both of which fuse the query and video features using a co-attention layer.

The results are summarized in Tables 2 and 3.

Table 2. Performance comparisons on the ActivityNet-Captions dataset.

| | Feature | IoU@0.3 | IoU@0.5 | IoU@0.7 |
|---|---|---|---|---|
| MCN [12] (20 | VGG | 21.37 | 9.58 | |
| CTRL [13] | C3D | 28.70 | 14.00 | |
| ABLR [24] | C3D | 55.67 | 36.79 | |
| QSPN [26] | C3D | 45.30 | 27.70 | 13.60 |
| CBP [18] | C3D | 54.30 | 35.76 | 17.80 |
| SCDM [21] | C3D | 54.80 | 36.75 | 19.86 |
| CMIN [19] | C3D | **63.61** | 43.40 | 23.88 |
| 2D-TAN [23] | C3D | 59.45 | 44.51 | 26.54 |
| TMLGA [25] | I3D | 51.28 | 33.04 | 19.26 |
| LGI [36] | C3D | | 41.51 | 23.07 |
| PMI [16] | C3D | 59.69 | 38.28 | 17.83 |
| CPNet [17] | C3D | | 40.56 | 21.63 |
| TACI | C3D | 61.67 | **45.50** | **27.23** |

On ActivityNet-Captions, the TACI outperforms the state-of-the-art methods at R@1, as shown in Table 2. For R@1, the TACI outperforms the best and second state-of-the-art methods, 2D-TAN with 0.99%p and 0.69%p for IoU@0.5 and IoU@0.7, and CMIN with 1.11%p and 3.35%p for IoU@0.5 and IoU@0.7, respectively. In addition, the TACI achieves superior performances compared to recent state-of-the-art TML methods, such as LGI, PMI, and CPNet, which achieve lower performance than CMIN. On Charades-STA, the TACI outperforms state-of-the-art TML methods using VGG and C3D as the video features, as shown in Table 3. However, TACI achieves lower performance than TML methods using I3D features.

Note that ActivityNet-Captions is a better dataset for evaluating temporal moment localization than Charades-STA, as the former has more complex query sentences with human annotation, more queries per video and open-set videos containing various instances. The query sentences in Charades-STA are semi-automatically generated, and their length is limited to an average of 6 words per query as shown in Table 1. Videos are relatively short and is limited to indoor videos. The limitation of a query description with a few words can lead to ambiguous query sentences that match multiple moments in a whole video.

Table 3. Performance comparisons on the Charades-STA dataset

| Feature | | IoU@0.3 | IoU@0.5 | IoU@0.7 |
|---|---|---|---|---|
| VGG | MAN [20] | | 41.24 | 20.54 |
| | SAP [22] | | 27.42 | 13.36 |
| | 2D-TAN [23] | | 39.40 | 23.31 |
| | TACI | 60.86 | **43.53** | **23.46** |
| C3D | CTRL [13] | | 23.63 | 8.89 |
| | QSPN [26] | 54.70 | 35.60 | 15.80 |
| | CBP [18] | | 36.80 | 18.87 |
| | PMI [16] | | 39.73 | 19.27 |
| | CPNet [17] | | 40.32 | 22.47 |
| I3D | SCDM [21] | | 54.44 | 33.43 |
| | TMLGA [25] | 67.53 | 52.02 | 33.74 |
| | LGI [36] | | 59.46 | 35.48 |
| | CPNet [17] | | **60.27** | **38.74** |

Table 4. R@1, IoU@0.7 results for a query group that contains temporal transitions

| Temporal | TACI | 2D-TAN | CMIN |
|---|---|---|---|
| IoU@0.7 | **27.85** | 27.17 | 26.50 |

Table 5. R@1, IoU@0.7 results for a query group that contains spatial transitions

| Spatial | TACI | 2D-TAN | CMIN |
|---|---|---|---|
| IoU@0.7 | **31.29** | 28.64 | 24.26 |

Tables 4 and 5 show the in-depth comparisons for queries that contain temporal and spatial transitions with the two state-of-the-art methods, 2D-TAN and CMIN, on the performance in ActivityNet-Captions. A total of 1,517 temporal queries are selected that contain temporal transition words, such as *finally*, *afterwards*, *while*, *after*, *until*, and *before*. On temporal queries, the TACI obtains the best performance for R@1, IoU@0.7, which is 0.68%p and 1.35%p better than that of 2D-TAN and CMIN, respectively. 3,424 spatial queries are selected that contain spatial transition words, such as *above*, *across*, *around*, *behind*, *beside*, *between*, *inside*, *near*, *outside*, *over*, *under*, *front*, *left*, and *right*. On spatial queries, the TACI obtains the best performance for R@1, IoU@0.7, which is 2.65%p and 7.03%p better than that of 2D-TAN and CMIN, respectively. The results show that the TACI can better localize moments containing both temporal and spatial submoments.

For an efficiency analysis, we compare the execution times of the proposed method with 2D-TAN in Table 6. We measure the runtime as the total time needed to localize 17,031 test videos of ActivityNet-Captions using sentence queries. We set the batch size to 32 in all cases and exclude the time needed for the feature extraction of sentences by GloVe-300 and video by C3D. The tests are conducted in the same hardware conditions, using the Intel i7 10700, Nvidia GTX 1080ti, and 32GB RAM. Compared to 2D-TAN, the TACI has a significantly faster runtime. Compared to 2D-TAN, both studies are based on generating 2D proposal maps. However, 2D-TAN is based on multiple layers of a 2D temporal adjacent network composed of convolutional layers, which make the network slower. Our model is significantly faster than 2D-TAN and at the same time, shows a slightly higher performance.

Table 6. Comparison of the execution time with performance

| Performance | TACI | 2D-TAN |
|---|---|---|
| Runtime (seconds) | **26.51** | 862.10 |
| Performance (R@1, IoU@0.7) | **27.23** | 26.54 |

*4.5. Ablation Studies*

In this section, we evaluate various experiments to demonstrate the effectiveness of the proposed modules in the TACI. Specifically, we retrain our models with the following conditions:

- Non-Attention (NA): This removes the attention stream in the two-stream attention and uses the non-attention stream only.
- Single Attention (SA): This removes the non-attention stream in the two-stream attention and uses the attention stream only.
- Two-stream Attention (TA): This uses the proposed two-stream attention.
- Global Evaluation (GE): This removes the local evaluation submodule and uses the global evaluation submodule only.
- Local Evaluation (LE): This removes the global evaluation submodule and uses the local evaluation submodule only.
- Global and Local Evaluation (GL): This uses both the global and local evaluation submodules.

Table 7. R@1, IoU@0.7 evaluation results of the ablation models for the local-global evaluation module

| | ActivityNet-Captions | | Charades-STA | |
|---|---|---|---|---|
| | R@1, IoU@0.5 | R@1, IoU@0.7 | R@1, IoU@0.5 | R@1, IoU@0.7 |
| GE | 44.82 | 25.16 | 41.02 | 22.11 |
| LE | 44.13 | 25.26 | 40.35 | 21.71 |
| GL | **45.50** | **27.23** | **43.53** | **23.46** |

As shown in Table 7, the early attention layers for the query and video features show performance improvements in both the ActivityNet-Captions and Charades-STA datasets. In particular, the model adopting the proposed two-stream attention showed the best performance in all cases. Although the attention layer used in existing studies showed its effectiveness, the attention layers may suppress important clip features. The proposed two-stream attention structure that outputs both the attended and unattended features enables the TACI to learn each part of the contextual information individually.

As shown in Table 8, the TACI with both global and local evaluation submodules shows the best performance in both the ActivityNet-Captions and Charades-STA datasets. In addition, the model using only the local evaluation submodule shows better performance than that using only the global evaluation submodule, which is most widely adopted in existing methods for temporal moment localization. These results imply that the two evaluation submodules are complementary. The local evaluation submodule enables the TACI to understand the video in a boundary-aware way, which helps to distinguish the query-related actions from other actions. The global proposal evaluation submodule enables the model to better understand the contextual information of the action from the whole video.

Table 8. R@1, IoU@0.7 evaluation results of the ablation models for the two-stream attention module

| | ActivityNet-Captions | | Charades-STA | |
|---|---|---|---|---|
| | R@1, IoU@0.5 | R@1, IoU@0.7 | R@1, IoU@0.5 | R@1, IoU@0.7 |
| NA | 44.00 | 24.37 | 40.51 | 20.44 |
| SA | 44.89 | 25.84 | 43.13 | 22.68 |
| TA | **45.50** | **27.23** | **43.53** | **23.46** |

*4.6. Effectiveness of CM-LSTM*

In this section, we evaluate extensive experiments to demonstrate the effectiveness of our proposed cross-modal LSTM. In Table 9, we compare CM-LSTM and the other fusion schemes:

- Element-wise Multiplication (EM): We apply element-wise multiplication to the last output of the sentence Bi-LSTM and video features and then feed them into another Bi-LSTM to replace the cross-modal LSTM.
- Concatenation (CAT): We apply a concatenation with the last output of the sentence Bi-LSTM and video features and then feed them into another Bi-LSTM to replace the cross-modal LSTM.
- CTRL-like, (CTRL): Instead of cross-modal LSTM, we apply element-wise addition, element-wise multiplication, and fusion through a fully connected layer to the last output of the sentence Bi-LSTM and video features. This method was proposed in [13]. We concatenate these three features and feed them into another Bi-LSTM.

Table 9. Results for various modality fusion methods on the ActivityNet-Captions dataset

| | IoU@0.3 | IoU@0.5 | IoU@0.7 |
|---|---|---|---|
| EM | 60.45 | 43.16 | 24.37 |
| CTRL | 60.63 | 44.21 | 25.99 |
| CAT | 61.58 | 44.58 | 25.75 |
| CM-LSTM | 61.67 | **45.50** | **27.23** |

Table 9 shows the effectiveness of our fusion scheme using cross-modal LSTM. TACI-EM has the worst results. Comparing CTRL and CAT, CM-LSTM shows the best improvement for IoU@0.5 and IoU@0.7. In particular, CM-LSTM is more effective for high-quality localization results with high IoU values. This shows the contribution of CM-LSTM to the TACI for performance improvement.

To show the effectiveness of our proposed CM-LSTM when applied to other methods for temporal moment localization, we also measure the performance improvement gained by replacing the recurrent layer of the recent methods of temporal moment localization, ABLR, CMIN, and TMLGA with CM-LSTM. As shown in Table 10, all three methods with the recurrent layer replaced by CM-LSTM show enhanced performance across all IoU cases. We replaced the existing recurrent layers located after video feature extraction in the three methods with CM-LSTM. In ABLR, the performance improvement is 2.36%p and 1.28%p for IoU@0.3 and IoU@0.5, respectively. In CMIN, the performance improvement is 0.75%p, 2.61%p and 2.13%p for IoU@0.3, IoU@0.5, and IoU@0.5, respectively. In TMLGA, the performance improvement is 2.5%, 1.72%p, and 0.91%p for IoU@0.3, IoU@0.5, and IoU@0.7, respectively.

Table 10. Performance improvements of the three TML methods by adopting CM-LSTM

| Methods | | Feature | Dataset | IoU@0.3 | IoU@0.5 | IoU@0.7 |
|---|---|---|---|---|---|---|
| ABLR [24] | Original | C3D | ActivityNet-Captions | 55.67 | 36.79 | |
| | Replaced | | | 58.04 | 38.07 | |
| CMIN [19] | Original | C3D | ActivityNet-Captions | 63.31 | 43.40 | 23.88 |
| | Replaced | | | 64.06 | 46.01 | 26.01 |
| TMLGA [25] | Original | I3D | Charades-STA | 67.53 | 52.02 | 33.74 |
| | Replaced | | | 70.03 | 53.74 | 34.65 |

### 4.7. Qualitative Analysis

In Fig. 5, we present a qualitative evaluation of our temporal moment proposals on ActivityNet-Captions. Figure 5 shows the case of 3 different queries localizing different intervals in a video. The results show a better performance improvement for precise localization using more TACI modules.

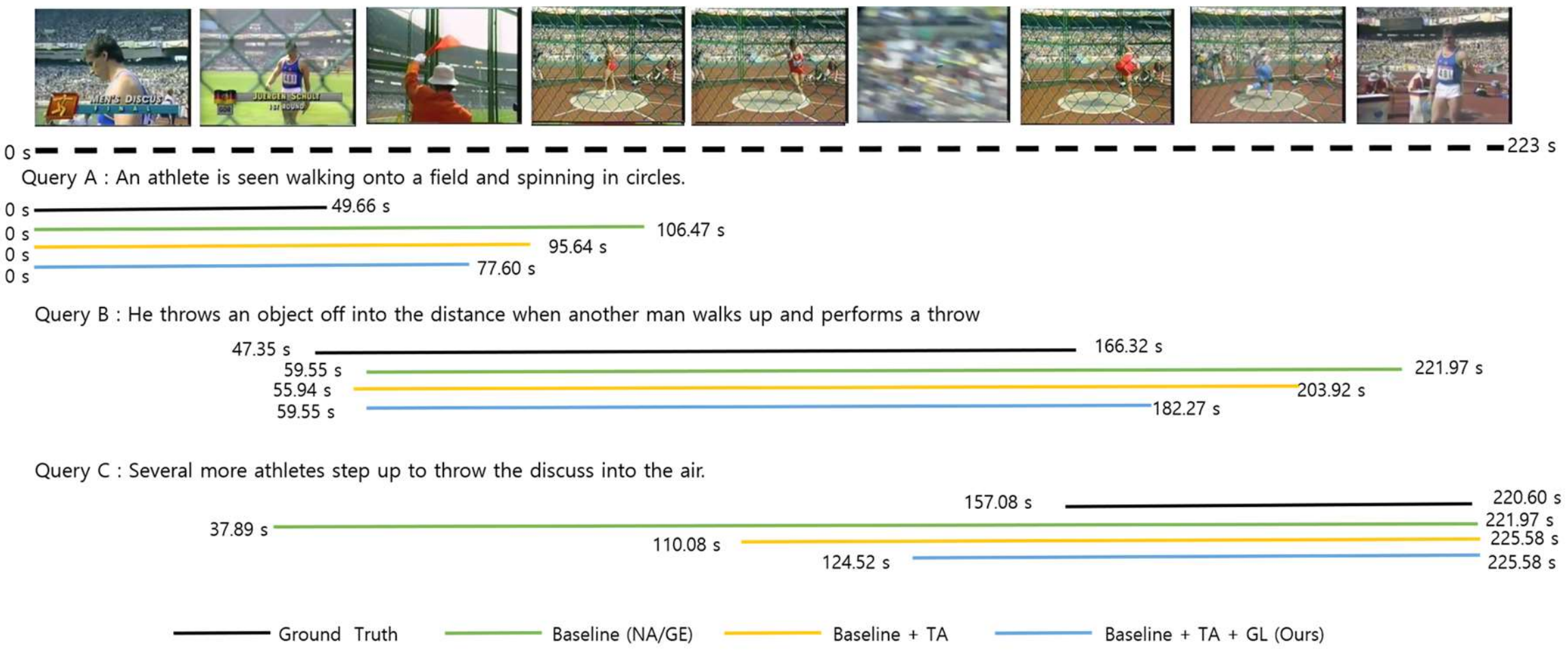

Fig. 5. Qualitative evaluation of successive cases with ActivityNet-Captions.

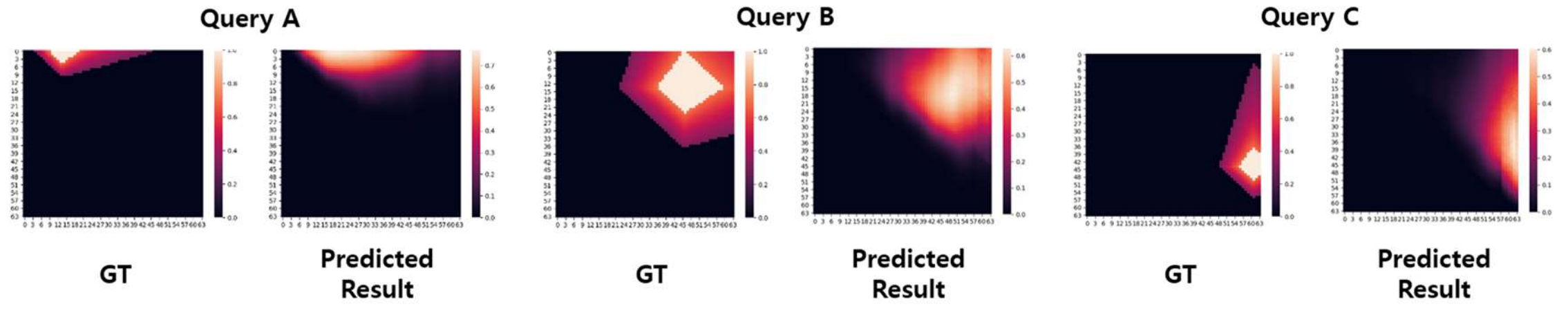

Fig. 6. Visualization of the 2D proposal map for queries A, B, and C.

The 2D map for temporal proposals, which is the final result of the TACI, is shown in Fig. 6. The left is the ground-truth map, and the right is the predicted map. The brightest value represents the best confidence score for the moment described by the input query. These maps show a tendency to focus on sections with specific actions and exclude sections that contain irrelevant moments.

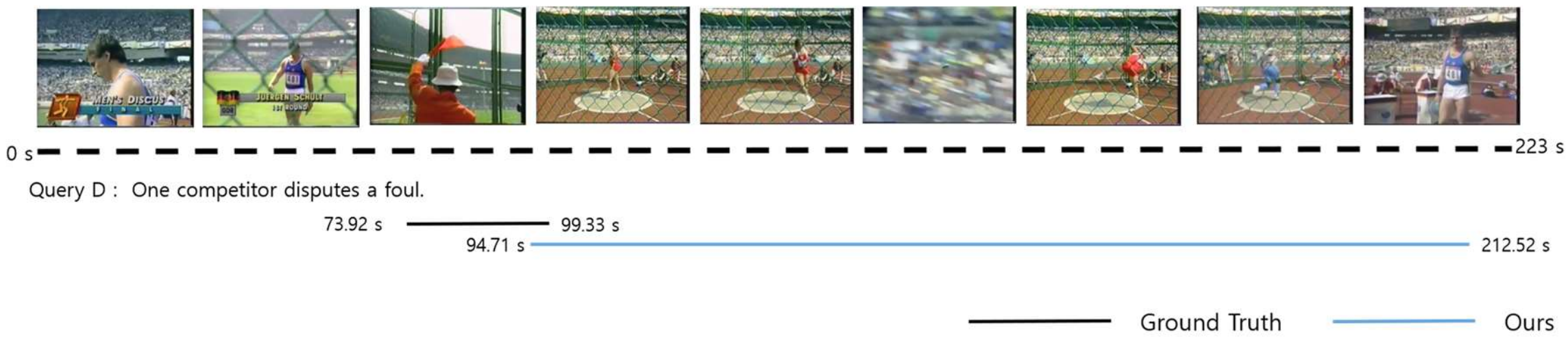

Fig. 7. Qualitative evaluation of a failure case with ActivityNet-Captions.

Fig. 7 shows how our model fails on query D, "One competitor disputes a foul". The action of disputing a foul is complex and abstract. To successfully localize this query, we should understand the actions of fouling, a competitor's disputing, and the referee's giving out a red card. The TACI localizes actions in which a second athlete appears. The localization result obtained by the TACI is the part after the second athlete appears. We can infer that in query D, *one competitor* was partially understood. We assume that disputing a foul was not learned well. For more accurate learning, more moment-sentence pairs are required.

## 5. Conclusion

In this paper, we present a novel end-to-end architecture that is designed to temporally localize a moment described by a complex sentence query with precise boundaries in a video. First, we introduce the CM-LSTM unit to mimic the human cognitive process for temporal moment localization, which compares a video segment with the query and accumulates contextual information throughout the entire video. Existing methods fuse independently generated video-level and query embeddings by using simple vector operations, self-sided modal attention, or cross-modal attention; in contrast, the TACI applies cross-modal attention to each video segment and to the query using the proposed CM-LSTM, which integrates contextual information recurrently by considering the modal relationship between the parts of a video segment and the query. This enables the TACI to focus on only the visual embeddings related to the query embeddings in a frame-wise way. Second, we generate a 2D proposal map by combining the two 2D proposal maps obtained globally from the integrated contextual features and the three predicted boundary-aware score sequences obtained locally for the start, end, and momentness, respectively, in an end-to-end manner. Most methods evaluate moment proposals globally, and some methods generate moment proposals heuristically using the predicted boundary scores.

It is expected that applying CM-LSTM can also be beneficial in both performance and speed in other tasks that require understanding complex relationships between two different modalities, such as video captioning and video QA, as applying CM-LSTM improves the performance of existing methods, such as ABLR [24], CMIN [19], and TMLGA [25], with faster execution time.

There are still issues to be considered for localizing temporal moments in a video. First, existing methods, including the TACI, use the fixed number of moment candidates in either a proposal map or an anchor set. Thus, when they take a long video, which is compared to videos used for training, as input, they fail to capture precise ground-truth moments with the fixed number of moment candidates. Second, existing networks, including the BSN, are trained and tested by using closed-set temporal moment localization datasets assuming that a query specifying a moment or its similar queries exists in the training set. This assumption is impractical because real-world videos contain moments with various situations and contexts.

In future work, we plan to design a model that adaptively generates video proposals that are suitable for variable lengths of videos and propose a mechanism for applying the models on an open-set dataset.

## Acknowledgements

This work was supported by Institute of Information & Communications Technology Planning & Evaluation (IITP) grant funded by the Korea government (MSIT) (No. 2020-0-00004, Development of Previsional Intelligence based on Long-term Visual Memory Network and No. 2014-3-00123, Development of High Performance Visual BigData Discovery Platform for Large-Scale Realtime Data Analysis)